\documentclass[11pt, a4paper, table]{article}

\usepackage[utf8]{inputenc}
\usepackage[T1]{fontenc}
\usepackage{amsmath}
\usepackage{newtxtext, newtxmath}
\usepackage{microtype}
\usepackage[left=1in, right=1in, top=1.2in, bottom=1.2in]{geometry}
\usepackage{setspace}
\usepackage{graphicx}
\usepackage{booktabs}
\usepackage{caption}
\usepackage[flushleft]{threeparttable} 
\usepackage{xcolor}
\usepackage{makecell}
\usepackage{tabularx}
\newcolumntype{Y}{>{\centering\arraybackslash}X}
\newcommand{\cellleft}[1]{%
  \ensuremath{\vcenter{\hbox{\setlength{\lineskip}{2.5pt}\shortstack[l]{#1}}}}%
}
\newcommand{\cellcenter}[1]{%
  \ensuremath{\vcenter{\hbox{\setlength{\lineskip}{2.5pt}\shortstack[c]{#1}}}}%
}
\usepackage{xcolor}
\usepackage{eso-pic}

\usepackage{authblk}
\usepackage[colorlinks=true, linkcolor=black, citecolor=blue, urlcolor=blue]{hyperref}

\DeclareCaptionLabelSeparator{pipe}{ \textbf{|} }
\newcommand{\doublerule}{\par\noindent\vbox{\hrule width\textwidth height 0.6pt \vskip 2pt \hrule width\textwidth height 0.6pt}\par}

\title{\vspace{-1.5cm}\Large \bfseries KneeFS-Net: A Lightweight Deep Learning Model for Knee Osteoarthritis Severity Assessment with Cross-Domain External Validation}

\author[1]{Shafqat Alam}
\author[1]{Galib Anjum Talukder Mahi}
\author[2]{Yeasir Rabbani Tanvir}
\author[3]{Md. Ibrahim}
\author[4]{Md. Shahed Hossain}
\author[4]{Shashank Ghimire}
\author[4]{Mahmud Hasan Mostofa Kamal}
\author[4]{ASM Shahidul Hossain}
\author[1,5]{Taufiq Hasan}

\affil[1]{\small Department of Biomedical Engineering, Bangladesh University of Engineering and Technology (BUET), Dhaka, Bangladesh}

\affil[2]{\small Telerad Bangladesh Ltd., Dhaka, Bangladesh}

\affil[3]{\small Department of Radiology and Imaging, DNCC Dedicated COVID 19 Hospital, Dhaka, Bangladesh}

\affil[4]{\small Department of Radiology and Imaging, Bangladesh Medical University, Dhaka, Bangladesh}

\affil[5]{\small Center for Bioengineering Innovation and Design, Department of Biomedical Engineering, Johns Hopkins University, Baltimore, MD, USA}

\date{}

\begin{document}

\maketitle

\vspace{-1cm}
\doublerule
\vspace{-0.5cm}
\section*{Abstract}
Knee osteoarthritis (OA) grading from radiographs using the Kellgren--Lawrence (KL) scale requires expertise that may be scarce. We present KneeFS-Net, a lightweight ordinal grader (0.8 million parameters; 0.33 GFLOPs) combining spatial, Fourier and state-space features with landmark-guided attention. On the Osteoarthritis Initiative (OAI) test set, KneeFS-Net achieved a quadratic weighted kappa (QWK) of 0.863, mean absolute error (MAE) of 0.325 and 97.2\% within-one-grade accuracy, comparable to models with 9--35-fold more parameters. We evaluated cross-population generalization on two Bangladeshi cohorts, a single-center hospital cohort and a multi-center teleradiology cohort. Without further training or adaptation, the OAI-trained model achieved QWK 0.803, MAE 0.339 and 98.2\% within-one-grade accuracy on its hospital cohort, declining to QWK 0.656 and MAE 0.522 on multi-center teleradiology images. Self-supervised pretraining on 6,092 unlabeled radiographs followed by joint fine-tuning achieved QWK 0.823, MAE 0.329 and 99.7\% within-one-grade accuracy in cross-validation. These findings demonstrate accurate, efficient KL grading and potential adaptation to new populations with limited annotation.
\doublerule

\section*{Introduction}
\label{Introduction}

Knee osteoarthritis (KOA) is a significant cause of pain and disability, and its impact is predicted to rise with population aging and the increasing prevalence of risk factors including excess body weight \cite{hunter_osteoarthritis_2019, gbd_global_2023}. In 2020, osteoarthritis affected an estimated 595 million people worldwide, with the knee being the most commonly affected joint \cite{gbd_global_2023}. This burden is especially significant in South Asia, including Bangladesh. In a nationally representative cross-sectional study of 1,843 adults in Bangladesh, 7.3\% were identified as having KOA, and the prevalence rose from 1.5\% in those aged 18--37 years to 14.8\% in those aged $\geq$ 58 years. Subjects with KOA were also more likely to report work loss \cite{haider_risk_2022}. In India, population-based studies have found a high prevalence of KOA, with a recent meta-analysis of 16 community-based studies estimating a pooled prevalence of 20.24\% among adults, although prevalence estimates varied considerably across studies and diagnostic methods used in the region \cite{hazra_prevalence_2025}. In a further meta-analysis of 14 studies, the pooled prevalence was estimated at 47.0\% among older adults in India, further underscoring the relevance of KOA in aging South Asian populations \cite{daniel_prevalence_2025}. These data collectively suggest a substantial and possibly increasing burden of KOA in the region and highlight the need for scalable approaches to structural assessment and disease monitoring.

Plain radiography is the main imaging technique for structural evaluation of KOA, with the Kellgren--Lawrence (KL) grading system offering a five-grade classification framework based on features including osteophytes, joint-space narrowing, and bony changes \cite{kellgren_radiological_1957}. However, KL grading is semi-quantitative and depends on reader interpretation, and its known limitations include variability between readers, particularly in distinguishing adjacent early grades \cite{kohn_classifications_2016}. Early computer-aided methods tried to quantify these features explicitly, such as joint-space width, osteophyte formation, and alignment, but required substantial feature engineering and anatomical localization \cite{shamir_knee_2009}. In recent years, the advent of deep learning has shifted this toward automatic representation learning directly from X-rays \cite{antony_quantifying_2016}. Tiulpin et al. showed that convolutional neural networks (CNNs) were capable of automatically detecting and grading KOA from plain radiographs, paving the way for subsequent data-driven methods \cite{tiulpin_automatic_2018}. Later research recognized that KL grading is an ordinal task and proposed specific ordinal-learning objectives to address the closer similarity between adjacent KL grades compared with those further apart \cite{Chen2019KneeOAOrdinalLoss}. Further studies added more information, including radiographic features, anatomical localization, multiview information, and prior regional knowledge to enhance grading performance, marking a shift from generic image classification toward models that explicitly capture clinically relevant joint structures \cite{tiulpin_automatic_2020, li_deep_2023, li_integrating_2024}.

As the use of deep learning to assess KOA continued to progress, studies started to go beyond automated KL classification to improve the clinical relevance and robustness of the assessment \cite{leung_prediction_2020}. It has been found that deep neural networks can achieve agreement with radiologists when grading severity from radiographs, while approaches using feature learning from radiographs and attention have sought to recognize clinically relevant features such as osteophytes and joint-space narrowing \cite{thomas_automated_2020, swiecicki_deep_2021}. Anatomical localization and region-specific feature modeling have also been adopted to guide the network toward clinically relevant areas of the knee instead of relying on whole-image representations \cite{tiulpin_kneel_2019, wang_automatic_2021, lee_accurate_2024}. Later research has also started to test models on independent clinical datasets in addition to randomly held-out images from the development dataset \cite{brejnebol_external_2022, vaattovaara_kellgren-lawrence_2025}. However, a systematic review of deep learning algorithms for radiologic diagnosis found that most showed reduced performance on external datasets, indicating that performance in the development population may not transfer to new clinical settings \cite{yu_external_2022}.

Because target-domain data are often sparsely labeled, there is growing interest in methods that transfer knowledge across imaging populations without extensive additional annotation. Domain adaptation (DA) addresses this problem by accounting for distributional differences between a labeled source domain and a target domain, which may arise from differences in patient populations, acquisition protocols, scanners, or clinical settings \cite{guan_domain_2022}. In medical imaging, unsupervised domain adaptation (UDA), which uses labeled source data and unlabeled target data, has been studied through feature alignment, image translation, pseudo-labeling, and related approaches \cite{kumari_deep_2024}. A complementary approach is self-supervised learning (SSL), in which representations are learned from unlabeled images through pretext or contrastive tasks and then fine-tuned with task-specific supervision \cite{huang_self-supervised_2023, krishnan_self-supervised_2022}. Self-supervised pretraining has been shown to improve downstream medical imaging tasks, especially when labeled data are scarce, and pretraining on unlabeled images relevant to the target domain can further improve transferability \cite{azizi_big_2021, zhou_models_2021}. Large-scale self-supervised pretraining has also improved robustness to distribution shift and reduced the amount of labeled data required in new clinical settings \cite{azizi_robust_2023}. However, this advantage depends on the similarity between the pretraining and target domains, and a domain-specific self-supervised model may lose its benefit under distribution shift \cite{anton_how_2022}. Combining unlabeled radiographs with population-specific adaptation remains relatively under-investigated for KOA. Population differences are also relevant to radiographic KOA assessment itself: on knee radiographs, a deep learning-derived severity measure explained substantially more of the pain disparities in underserved patients than KL grades, which were developed in historically narrow populations \cite{pierson_algorithmic_2021}. This gap is especially relevant for South Asia, where the burden of KOA is substantial but large, real-world, expert-annotated radiographic datasets remain limited. Methods that leverage abundant unlabeled radiographs without extensive target-domain annotation could therefore provide a practical pathway toward population-specific KOA modeling.

In this work, we developed KneeFS-Net, a lightweight framework for ordinal KOA grading designed to address population shift and computational constraints in underrepresented clinical settings. The framework combines frequency-domain, spatial, and selective state-space feature learning with landmark-guided regional attention to capture global radiographic patterns and local joint abnormalities, while an ordinal prediction head explicitly models the ordered nature of KL grades. The framework was first developed using the large, publicly available OAI cohort and was then evaluated on BUET-KOA, a curated dataset of two Bangladeshi cohorts: the Bangladesh Medical University (BMU) Cohort, a single-center cohort in which all knees were independently annotated by three radiologists, and the Multi-District Cohort, a large, real-world teleradiology cohort that included unlabeled clinical images and a retrospectively annotated subset, in which 1,000 knees were independently annotated by the same three-radiologist protocol. This design allowed us to assess generalization across different clinical environments and investigate whether population-specific unlabeled radiographs could support adaptation to a previously underrepresented target population. KneeFS-Net has approximately 0.8 million trainable parameters and was designed to maintain clinically relevant predictive performance while requiring fewer computational resources than high-capacity deep-learning architectures.

\section*{Results}

\subsection*{In-domain performance and computational efficiency}

\begin{figure}[!t]
    \centering
    \includegraphics[width=\textwidth]{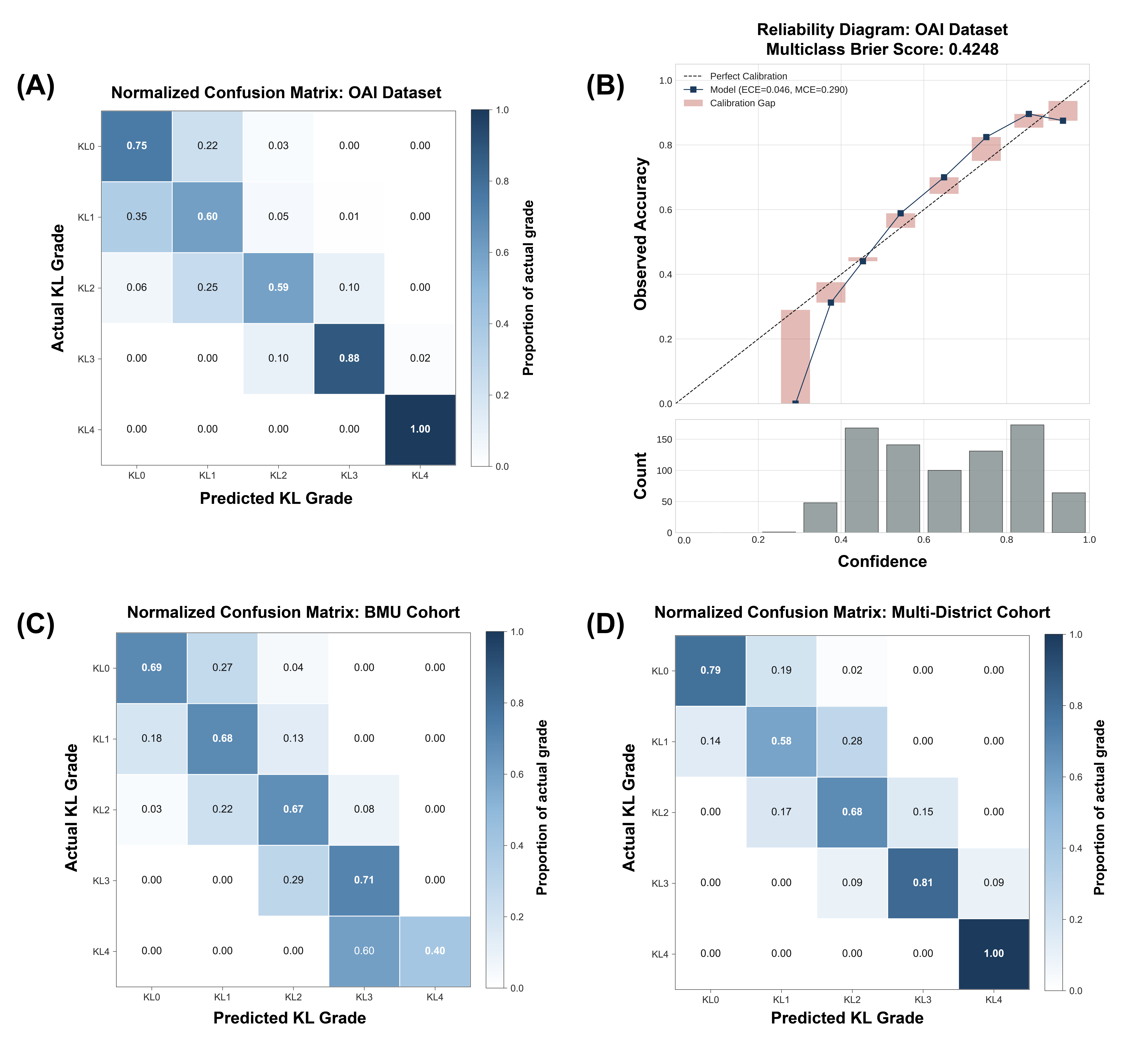}
    \caption{\textbf{Grading errors in the three cohorts and calibration on OAI.} \textbf{A} OAI test set ($n = 826$). \textbf{B} Reliability diagram on the OAI test set; bars show the number of knees per confidence bin. \textbf{C} BMU Cohort ($n = 271$). \textbf{D} Multi-District Cohort, Experiment F, out-of-fold predictions ($n = 1{,}000$). Confusion matrices are row-normalized, so the diagonal gives the sensitivity for each grade.}
    \label{fig:performance}
\end{figure}

KneeFS-Net graded the OAI test set as well as networks 9 to 35 times its size, with 0.8 million parameters. On the 826 held-out knees (see Methods, \hyperref[oai]{OAI}), it reached a quadratic weighted kappa (QWK) of 0.863 (95\% CI, 0.840--0.883) against the reference grades and graded 70.3\% of knees exactly. Only 23 knees (2.8\%) were misgraded by more than one grade, and the model slightly more often under-graded (14.4\%) than over-graded (12.5\%). Errors were concentrated at the mild end of the scale: 35\% of KL1 knees were called KL0 and 25\% of KL2 knees were called KL1, which made KL1 the hardest grade (sensitivity 59.9\%, precision 41.6\%; {Fig.~\ref{fig:performance}A}). Severe disease was graded reliably, and every KL4 knee in the test set was identified. Predicted probabilities matched observed accuracy closely (expected calibration error 0.046, multiclass Brier score 0.425; {Fig.~\ref{fig:performance}B}).

\begin{figure}[!t]
    \centering
    \includegraphics[width=\textwidth]{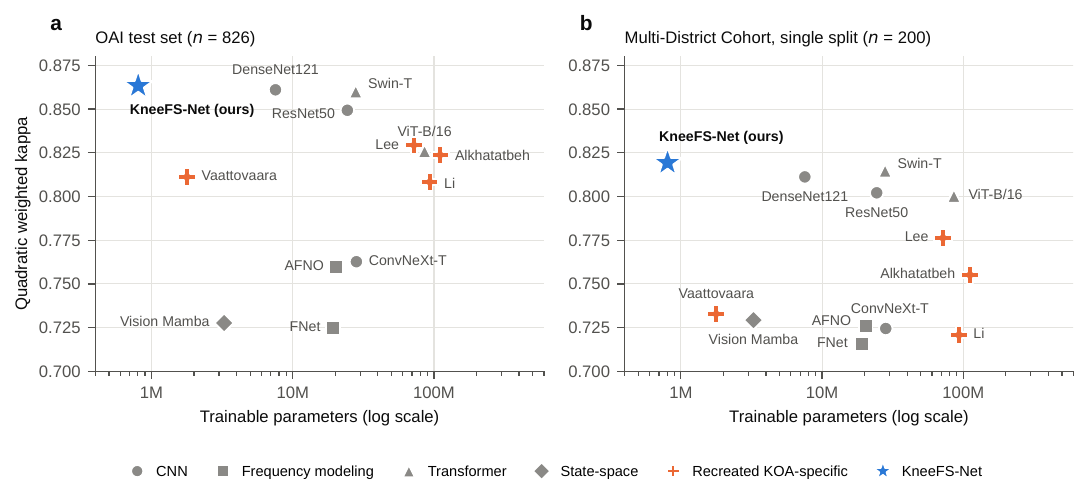}
       \caption{\textbf{Agreement versus model size.} QWK against the number of trainable parameters (log scale). \textbf{A} OAI test set ($n = 826$). \textbf{B} Multi-District Cohort, single 200-knee split. Recreated KOA-specific models are labeled by first author; full results are in Table~\ref{tab:benchmark}.}
    \label{fig:params}
\end{figure}

We compared KneeFS-Net with eight established CNN, frequency-modeling, Transformer and state-space architectures and with four recently published KOA-specific models, which we recreated from their descriptions (see Methods, \hyperref[comparators]{Comparator models}). All models were trained and tested on the same splits (Table~\ref{tab:benchmark}). KneeFS-Net matched the best of these on OAI, DenseNet121 (QWK 0.861) and Swin-T (0.860), while using 9.4 and 34.7 times fewer parameters (Fig.~\ref{fig:params}A), and it had the highest macro-sensitivity and macro-specificity (Table~\ref{tab:benchmark}). It was also the least costly model to run: it needed 0.33 GFLOPs per image, about 18 times fewer than DenseNet121 (5.79) and Swin-T (5.99), and less peak memory than any established architecture (222 MB, against 435 and 682 MB). Its latency, however, was higher (9.5 vs.\ 3.1 and 4.5 ms per image), so its advantage lies in computation and memory rather than speed. The two other models with fewer than 4 million parameters, the recreated model of Vaattovaara et al.\ and Vision Mamba, reached QWKs of 0.811 and 0.728. The recreated KOA-specific models reached 0.808--0.829 with 2.2 to 138 times more parameters than KneeFS-Net. On the Multi-District Cohort, after the same target-domain adaptation was applied to the established architectures (Fig.~\ref{fig:params}b), KneeFS-Net, Swin-T and DenseNet121 reached similar QWK (0.819, 0.815 and 0.811), differences well within the uncertainty of a 200-knee test split, and five of the established architectures graded more knees exactly. This comparison used a single split rather than cross-validation.

\begin{table}[!b]
\centering
\sffamily
\caption{Performance and computational cost of KneeFS-Net and comparator models}
\label{tab:benchmark}
\setlength{\aboverulesep}{0pt}
\setlength{\belowrulesep}{0pt}
\setlength{\tabcolsep}{4pt}
\renewcommand{\arraystretch}{1.2}
\resizebox{\textwidth}{!}{%
\begin{tabular}{l rcc cccc cccc}
\toprule[1.5pt]
\rowcolor{black!10}
\rule{0pt}{1.3em} & \multicolumn{3}{c}{\textbf{Model size and cost}} & \multicolumn{4}{c}{\textbf{OAI test set ($n = 826$)}} & \multicolumn{4}{c}{\textbf{Multi-District, single split ($n = 200$)}} \\
\cmidrule(lr){2-4}\cmidrule(lr){5-8}\cmidrule(lr){9-12}
\rowcolor{black!10}
\textbf{Model} &
\makecell[c]{\textbf{Params} $\downarrow$} &
\makecell[c]{\textbf{GFLOPs}\\\textbf{(per img)} $\downarrow$} &
\makecell[c]{\textbf{Peak Mem.}\\\textbf{(MB)} $\downarrow$} &
\makecell[c]{\textbf{QWK} $\uparrow$} &
\makecell[c]{\textbf{Acc} $\uparrow$} &
\makecell[c]{\textbf{Sens.}\\\textbf{(macro)} $\uparrow$} &
\makecell[c]{\textbf{Spec.}\\\textbf{(macro)} $\uparrow$} &
\makecell[c]{\textbf{QWK} $\uparrow$} &
\makecell[c]{\textbf{Acc} $\uparrow$} &
\makecell[c]{\textbf{Sens.}\\\textbf{(macro)} $\uparrow$} &
\makecell[c]{\textbf{Spec.}\\\textbf{(macro)} $\uparrow$} \\[0.5ex]
\midrule[1pt]
\multicolumn{12}{l}{\textbf{Part A: Established architectures}} \\
\midrule
\multicolumn{12}{l}{\textit{\textbf{CNN-based Models}}} \\
ResNet50     & 24,363,588 & 8.26  & 840.71  & 0.8493 & 68.16 & 72.81 & 91.39 & 0.8021 & 71.50 & 74.85 & 90.22 \\
DenseNet121  & 7,547,268  & 5.79  & 434.59  & 0.8610 & \textbf{70.80} & 74.10 & 91.35 & 0.8112 & 71.50 & 75.21 & 93.27 \\
ConvNeXt-T   & 28,226,788 & 8.97  & 999.52  & 0.7627 & 56.42 & 60.72 & 88.02 & 0.7245 & 67.00 & 70.64 & 84.91 \\
\midrule
\multicolumn{12}{l}{\textit{\textbf{Frequency Modeling}}} \\
FNet         & 19,257,508 & 5.99  & 843.38  & 0.7248 & 47.94 & 54.52 & 86.45 & 0.7157 & 65.00 & 65.31 & 82.68 \\
AFNO         & 20,353,444 & 5.96  & 835.96  & 0.7597 & 50.43 & 55.05 & 87.41 & 0.7258 & 67.50 & 72.07 & 85.51 \\
\midrule
\multicolumn{12}{l}{\textit{\textbf{Transformer-based}}} \\
ViT-B/16     & 85,787,652 & 22.46 & 1170.40 & 0.8259 & 63.68 & 70.51 & 90.25 & 0.8003 & \textbf{72.50} & 74.53 & 91.32 \\
Swin-T       & 27,926,014 & 5.99  & 681.62  & 0.8600 & 69.07 & 72.62 & 91.06 & 0.8147 & 71.00 & 76.43 & \textbf{93.63} \\
\midrule
\multicolumn{12}{l}{\textit{\textbf{State-space Modeling}}} \\
Vision Mamba & 3,265,706  & 0.96  & 672.04  & 0.7276 & 50.36 & 57.12 & 86.61 & 0.7293 & 70.00 & 67.09 & 83.13 \\
\midrule[1pt]
\multicolumn{12}{l}{\textbf{Part B: Recreated KOA-specific models}} \\
\midrule
Lee et al.~\cite{lee_accurate_2024} & 71,815,032 & \--- & \--- & 0.8292 & 66.71 & 67.21 & 90.50 & 0.7761 & 65.50 & 66.19 & 87.41 \\
Vaattovaara et al.~\cite{vaattovaara_kellgren-lawrence_2025} & 1,785,231 & \--- & \--- & 0.8111 & 65.74 & 70.73 & 90.69 & 0.7328 & 65.00 & 69.05 & 85.45 \\
Li et al.~\cite{liAGENetSpectralSpatialFusion2026} & 93,040,774 & \--- & \--- & 0.8081 & 62.23 & 59.66 & 89.71 & 0.7208 & 63.00 & 66.53 & 83.28 \\
Alkhatatbeh et al.~\cite{alkhatatbehCrossInstitutionalFiveClassKellgren2026} & 111,125,709 & \--- & \--- & 0.8237 & 65.25 & 64.48 & 90.29 & 0.7553 & 66.50 & 68.16 & 88.30 \\
\midrule[1pt]
\rowcolor{black!5}
\textbf{KneeFS-Net (Ours)} & \textbf{804,872} & \textbf{0.33} & \textbf{222.11} & \textbf{0.8633} & 70.34 & \textbf{76.29} & \textbf{92.02} & \textbf{0.8193} & 67.50 & \textbf{78.04} & 90.76 \\[0.5ex]
\bottomrule[1.5pt]
\end{tabular}%
}

\vspace{1ex}
\parbox{\textwidth}{
    \footnotesize \textnormal{\textit{Note:} Bold, best value in each column. Acc, Sens.\ and Spec.\ are percentages; Sens.\ and Spec.\ are macro-averaged. GFLOPs, billions of floating-point operations per image; Peak Mem., peak GPU memory; --, not measured. Part B models were re-implemented from their publications (see Methods). Multi-District values come from a single 200-knee split and therefore differ from Table~\ref{tab:telerad_results}.}
}
\end{table}

In an ablation on the OAI test set, removing the frequency mixer, the selective state-space branch, the landmark-region attention or the multiscale design each lowered QWK by 0.015--0.020, and removing the first three together lowered it by 0.030. Adding capacity did not help: the Large variant, with 1.8 times as many parameters, reached 0.862, whereas the Small variant, with 0.34 million parameters, still reached 0.844.

\subsection*{Performance under domain shift}

Without any retraining, the OAI model kept most of its agreement on radiographs from a Bangladeshi hospital, but not on radiographs collected through a teleradiology network. On the 271 knees of the BMU Cohort, its QWK against the three-radiologist consensus was 0.803 (95\% CI, 0.749--0.848), and only 5 knees were misgraded by more than one grade (Fig.~\ref{fig:performance}C; Table~\ref{tab:reader}). On the 1,000 graded knees of the Multi-District Cohort (Experiment A), QWK fell to 0.656 (0.617--0.694), and 51 knees were misgraded by more than one grade (Fig.~\ref{fig:md_forest}; Table~\ref{tab:telerad_results}).

Both cohorts differed from OAI in grade distribution: KL1, rather than KL0, was the most common grade in each. The difference between them was in image quality. On image-noise descriptors, BMU radiographs clustered with OAI, whereas Multi-District radiographs spread much more widely (Fig.~\ref{fig:cohorts}).

\subsection*{Effect of self-supervised pretraining and joint fine-tuning}

\begin{figure}[!b]
    \centering
    \includegraphics[width=\textwidth]{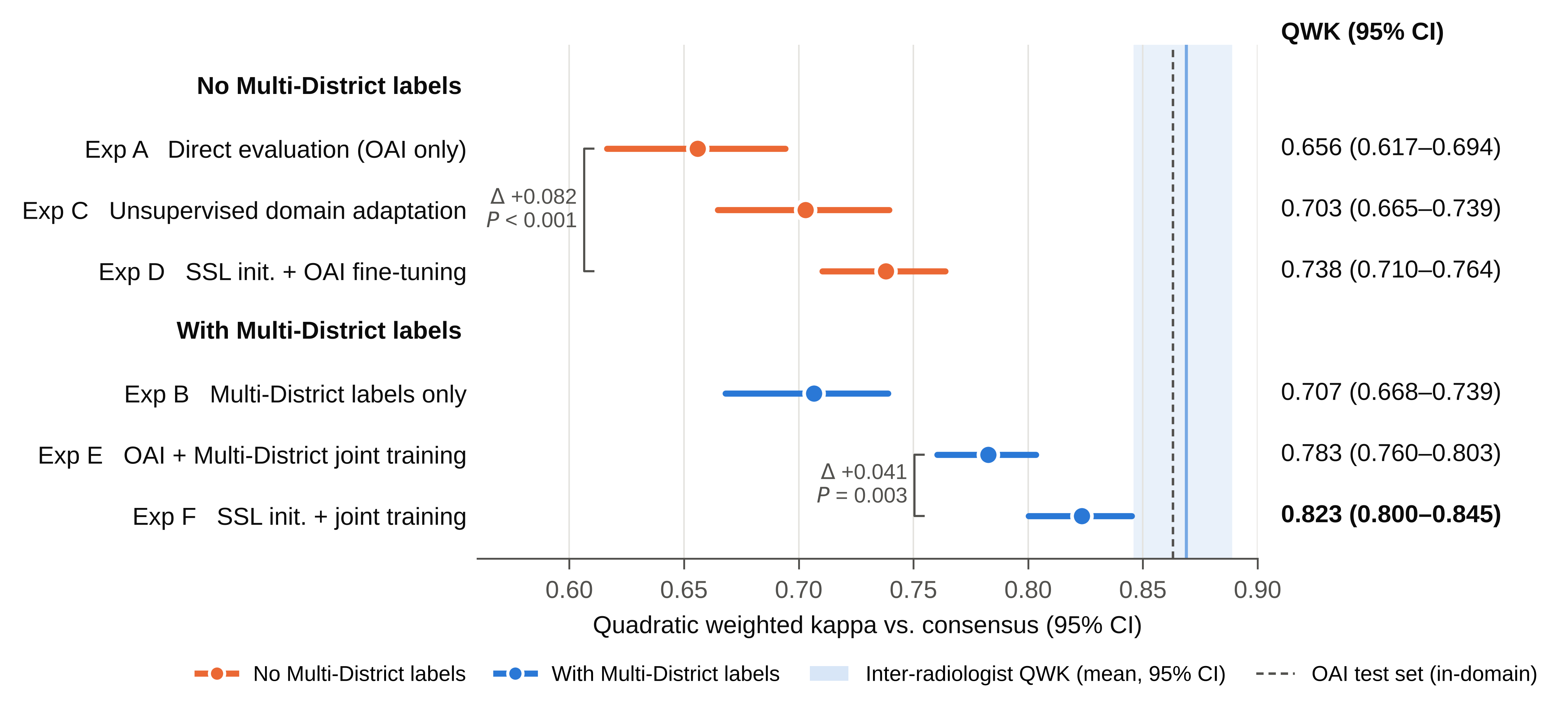}
    \caption{\textbf{Agreement on the Multi-District Cohort under six training strategies.} QWK against the consensus on out-of-fold predictions ($n = 1{,}000$), with 95\% CIs. Orange, no Multi-District labels used; blue, Multi-District labels used. The shaded band shows the mean inter-radiologist QWK (95\% CI) for context only, and the dashed line shows KneeFS-Net on the OAI test set. Brackets mark the planned comparisons (paired bootstrap tests).}
    \label{fig:md_forest}
\end{figure}

Combining self-supervised pretraining on unlabeled archive images with joint fine-tuning on a limited number of graded knees raised agreement on the Multi-District Cohort from 0.656 to 0.823 {($\Delta$QWK $= +0.167$; 95\% CI, 0.130--0.205; Fig.~\ref{fig:md_forest})}. We evaluated six strategies with 5-fold cross-validation over the 1,000 graded knees, so that every knee was graded by a model that had not seen it during training (see Methods, \hyperref[training]{Supervised training and adaptation experiments}).

\begin{table}[!t]
\centering
\sffamily
\begin{threeparttable}
\caption{Performance of six training strategies on the Multi-District Cohort (5-fold cross-validation, $n = 1{,}000$)}
\label{tab:telerad_results}
\setlength{\aboverulesep}{0pt}
\setlength{\belowrulesep}{0pt}
\renewcommand{\arraystretch}{1.3}
\setlength{\tabcolsep}{1.5pt}
\small

\begin{tabular}{l c c c c c c}
\toprule[1.5pt]
\rowcolor{black!10}
\rule{0pt}{1.3em}\textbf{Setup/ Metric} &
\textbf{Exp A} &
\textbf{Exp B} &
\textbf{Exp C} &
\textbf{Exp D} &
\textbf{Exp E} &
\textbf{Exp F} \\[0.5ex]
\midrule[1pt]

\rowcolor{gray!15}
\rule{0pt}{1.8em}\cellleft{\textbf{Exp. Name}} &
\cellcenter{Direct \\ Evaluation} &
\cellcenter{Trained on \\ MD} &
\cellcenter{UDA} &
\cellcenter{SSL init. + \\ OAI fine tune} &
\cellcenter{OAI + MD \\ Joint train} &
\cellcenter{SSL init. + \\ Joint train}\rule[-1.4em]{0pt}{0pt} \\
\midrule[1pt]

\rowcolor{gray!15}
\rule{0pt}{2.6em}\cellleft{\textbf{Train Data}} &
\cellcenter{OAI \\ Train set} &
\cellcenter{MD training \\ folds} &
\cellcenter{OAI Train + \\ MD unlabeled \\ ($n$=6,092)} &
\cellcenter{MD unlab. \\ ($n$=6,092) + OAI \\ Train} &
\cellcenter{OAI Train +\\ MD training \\ folds} &
\cellcenter{OAI Train +\\ MD training \\ folds}\rule[-1.4em]{0pt}{0pt} \\
\midrule[1pt]

\rule{0pt}{1.2em}\textbf{QWK}
& 0.6560 & 0.7067 & 0.7030 & 0.7379\textsuperscript{$***$} & 0.7826 & 0.8234\textsuperscript{$\dagger\dagger\ddagger\ddagger\ddagger$} \\
& \scriptsize (0.6165--0.6942) & \scriptsize (0.6681--0.7389) & \scriptsize (0.6648--0.7394) & \scriptsize (0.7103--0.7639) & \scriptsize (0.7604--0.8034) & \scriptsize (0.8002--0.8451) \\[0.4ex]

\textbf{Accuracy}
& 53.60 & 58.50 & 66.30 & 61.30\textsuperscript{$***$} & 61.90 & 67.40\textsuperscript{$\dagger\dagger$} \\
& \scriptsize (50.30--56.80) & \scriptsize (55.40--61.60) & \scriptsize (63.40--69.10) & \scriptsize (58.40--64.30) & \scriptsize (58.90--64.90) & \scriptsize (64.50--70.30) \\[0.4ex]

\textbf{Macro-F1}
& 50.37 & 46.02 & 52.00 & 53.32 & 50.04 & 71.49\textsuperscript{$\dagger\dagger\dagger\ddagger\ddagger\ddagger$} \\
& \scriptsize (45.71--54.67) & \scriptsize (43.05--48.77) & \scriptsize (49.17--54.74) & \scriptsize (49.52--57.11) & \scriptsize (47.35--52.56) & \scriptsize (67.90--74.86) \\[0.4ex]

\textbf{MAE}
& 0.5220 & 0.4380 & 0.3710 & 0.4390\textsuperscript{$***$} & 0.3860 & 0.3290\textsuperscript{$\dagger\dagger$} \\
& \scriptsize (0.4840--0.5610) & \scriptsize (0.4040--0.4740) & \scriptsize (0.3370--0.4050) & \scriptsize (0.4030--0.4750) & \scriptsize (0.3560--0.4170) & \scriptsize (0.2990--0.3580) \\[0.4ex]

\makecell[l]{\textbf{Within $\pm$}1 \\ \textbf{grade}}
& 94.90 & 98.30 & 97.10 & 94.80 & 99.50 & 99.70\textsuperscript{$\ddagger\ddagger\ddagger$} \\
& \scriptsize (93.50--96.20) & \scriptsize (97.50--99.00) & \scriptsize (96.00--98.10) & \scriptsize (93.40--96.10) & \scriptsize (99.00--99.90) & \scriptsize (99.30--100.00) \\[0.4ex]

\makecell[l]{\textbf{Macro} \\ \textbf{sensitivity}}
& 53.21 & 47.02 & 52.14 & 59.82\textsuperscript{$**$} & 53.63 & 77.26\textsuperscript{$\dagger\dagger\dagger\ddagger\ddagger\ddagger$} \\
& \scriptsize (48.03--58.57) & \scriptsize (43.97--50.22) & \scriptsize (49.23--55.04) & \scriptsize (55.93--64.19) & \scriptsize (50.91--56.53) & \scriptsize (74.76--79.89) \\[0.4ex]

\makecell[l]{\textbf{Macro} \\ \textbf{specificity}}
& 87.03 & 87.83 & 89.97 & 89.83\textsuperscript{$***$} & 89.23 & 90.77\textsuperscript{$\dagger$} \\
& \scriptsize (86.12--87.93) & \scriptsize (86.92--88.75) & \scriptsize (89.13--90.81) & \scriptsize (89.07--90.66) & \scriptsize (88.34--90.10) & \scriptsize (89.89--91.64) \\[0.5ex]

\bottomrule[1.5pt]
\end{tabular}

\begin{tablenotes}
    \item \footnotesize \textnormal{\textit{Note:} Values are computed on pooled out-of-fold predictions, with 95\% CIs below each estimate; accuracy, macro-F1, within $\pm$1 grade, macro-sensitivity and macro-specificity are percentages. MD, Multi-District; UDA, unsupervised domain adaptation; SSL, self-supervised learning. Paired bootstrap tests: $*$ versus Exp A (Exp D only), $\dagger$ versus Exp E and $\ddagger$ versus Exp C (Exp F only); one, two and three symbols denote $p < 0.05$, $p < 0.01$ and $p < 0.001$.}
\end{tablenotes}
\end{threeparttable}
\end{table}

Using Multi-District images without their labels helped only partly. Self-supervised pretraining followed by fine-tuning on OAI alone (Experiment D) improved QWK over direct transfer by 0.082 (95\% CI, 0.042--0.124; $p < 0.001$), to 0.738, and unsupervised domain adaptation (Experiment C) reached 0.703. A linear probe on the frozen self-supervised encoder reached a QWK of 0.573 on OAI validation, against 0.148 for a randomly initialized encoder, showing that pretraining without any grades had already captured features related to disease severity (see Methods, \hyperref[ssl]{Self-supervised target-domain pretraining}).

Graded Multi-District knees made the larger difference, but only in combination with OAI. Training on the Multi-District folds alone (Experiment B) gave a QWK of 0.707 and the lowest macro-F1 of all strategies (46.0\%), as the few KL3 and KL4 knees were not enough to learn these grades. Joint training on OAI and the Multi-District folds (Experiment E) raised QWK to 0.783, and initializing joint training from the self-supervised encoder (Experiment F) raised it further to 0.823 (95\% CI, 0.800--0.845). The gain from pretraining was modest in QWK ($+$0.041; 95\% CI, 0.016--0.067; $p = 0.003$) but large for the less frequent grades: macro-F1 rose from 50.0\% to 71.5\% and macro-sensitivity from 53.6\% to 77.3\% (both $p < 0.001$; Table~\ref{tab:telerad_results}). Experiment F was the best strategy on every metric, and results were stable across folds: the fold-to-fold standard deviation of QWK was at most 0.016 for any experiment and 0.006 for Experiment F.

In practical terms, the final model misgraded only 3 of the 1,000 knees by more than one grade, compared with 51 under direct transfer. All 23 KL4 knees were identified. Most remaining errors were at the mild end of the scale: KL1 knees accounted for 177 of the 326 misgraded knees, almost always assigned to a neighboring grade, and KL1 is also a grade on which readers often disagree \cite{kohn_classifications_2016}. KL3 was over-called, with 47 of the 108 knees predicted as KL3 assigned another grade by the consensus. Per-grade results are given in Fig.~\ref{fig:performance}D.

\begin{table}[!t]
\centering
\sffamily
\begin{threeparttable}
\caption{Agreement of KneeFS-Net with the radiologist consensus and with individual radiologists (BMU Cohort, direct evaluation; Multi-District Cohort, Experiment F)}
\label{tab:reader}
\setlength{\aboverulesep}{0pt}
\setlength{\belowrulesep}{0pt}
\renewcommand{\arraystretch}{1.25}
\setlength{\tabcolsep}{3pt}
\footnotesize
\begin{tabular}{l c c c c}
\toprule[1.5pt]
\rowcolor{black!10}
\rule{0pt}{1.3em} & \multicolumn{2}{c}{\textbf{BMU Cohort ($n = 271$)}} & \multicolumn{2}{c}{\textbf{Multi-District Cohort ($n = 1{,}000$)}} \\
\cmidrule(lr){2-3}\cmidrule(lr){4-5}
\rowcolor{black!10}
\textbf{Metric / Comparison} & \textbf{Estimate (95\% CI)} & \textbf{$\Delta$ (95\% CI)} & \textbf{Estimate (95\% CI)} & \textbf{$\Delta$ (95\% CI)} \\[0.5ex]
\midrule[1pt]
\multicolumn{5}{l}{\textit{\textbf{Overall model performance vs. consensus}}} \\
Exact accuracy (\%) & \makecell[c]{67.90\\(62.36--73.06)} & --- & \makecell[c]{67.40\\(64.50--70.30)} & --- \\
Within $\pm1$ grade (\%) & \makecell[c]{98.15\\(96.31--99.63)} & \makecell[c]{Differed $>1$ grade:\\1.85\% (5 knees)} & \makecell[c]{99.70\\(99.30--100.00)} & \makecell[c]{Differed $>1$ grade: \\0.30\% (3 knees)} \\
QWK & \makecell[c]{0.803\\(0.749--0.848)} & Primary metric & \makecell[c]{0.823\\(0.800--0.845)} & Primary metric \\
MAE & \makecell[c]{0.339\\(0.284--0.399)} & --- & \makecell[c]{0.329\\(0.299--0.358)} & --- \\
\midrule
\multicolumn{5}{l}{\textit{\textbf{Human--AI and inter-radiologist agreement (QWK)}}} \\
Model vs. consensus & \makecell[c]{0.803\\(0.749--0.848)} & Reference & \makecell[c]{0.823\\(0.800--0.845)} & Reference \\
Model vs. Radiologist A & \makecell[c]{0.765\\(0.699--0.818)} & \makecell[c]{$-0.101$\\($-0.152$ to $-0.050$)} & \makecell[c]{0.778\\(0.745--0.812)} & \makecell[c]{$-0.125$ \\($-0.152$ to $-0.099$)} \\
Model vs. Radiologist B & \makecell[c]{0.806\\(0.752--0.848)} & \makecell[c]{$-0.125$\\($-0.175$ to $-0.083$)} & \makecell[c]{0.751\\(0.718--0.780)} & \makecell[c]{$-0.104$ \\($-0.131$ to $-0.079$)} \\
Model vs. Radiologist C & \makecell[c]{0.780\\(0.723--0.825)} & \makecell[c]{$-0.126$\\($-0.178$ to $-0.077$)} & \makecell[c]{0.762\\(0.729--0.793)} & \makecell[c]{$-0.108$ \\($-0.129$ to $-0.085$)} \\
Mean inter-radiologist & \makecell[c]{0.849\\(0.813--0.881)} & \makecell[c]{$-0.046$\\($-0.093$ to $-0.002$)} & \makecell[c]{0.869\\(0.846--0.889)} & \makecell[c]{$-0.045$\\($-0.070$ to $-0.022$)} \\
Mean model--radiologist & \makecell[c]{0.784\\(0.747--0.819)} & \makecell[c]{$-0.065$\\($-0.112$ to $-0.018$)} & \makecell[c]{0.764\\(0.736--0.791)} & \makecell[c]{$-0.105$ \\($-0.125$ to $-0.084$)} \\[0.5ex]
\bottomrule[1.5pt]
\end{tabular}
\begin{tablenotes}
    \item \footnotesize \textnormal{\textit{Note:} MAE, mean absolute error. In the within $\pm$1 grade row, the $\Delta$ column gives the knees graded more than one grade from the consensus. In the Radiologist A--C rows, $\Delta$ is the model--consensus QWK minus that radiologist's QWK with the consensus, which is inflated because each radiologist contributed to the consensus. In the mean inter-radiologist row, $\Delta$ is the model--consensus QWK minus the mean inter-radiologist QWK; the two are not directly comparable. In the mean model--radiologist row (primary comparison), $\Delta$ is the mean model--radiologist QWK minus the mean inter-radiologist QWK; $p$ values for this comparison are given in the text.}
\end{tablenotes}
\end{threeparttable}
\end{table}

\subsection*{Agreement with radiologists}

In both Bangladeshi cohorts, the model agreed with individual radiologists less well than the radiologists agreed with each other (Table~\ref{tab:reader}). On the BMU Cohort, the mean model--radiologist QWK was 0.784, against 0.849 between radiologists ($\Delta$QWK $= -0.065$; 95\% CI, $-0.112$ to $-0.018$; $p < 0.05$). On the Multi-District Cohort, with Experiment F, the mean model--radiologist QWK was 0.764, against 0.869 between radiologists ($\Delta$QWK $= -0.105$; 95\% CI, $-0.125$ to $-0.084$; $p < 0.001$). The model's QWK with single radiologists ranged from 0.765 to 0.806 on the BMU Cohort and from 0.751 to 0.778 on the Multi-District Cohort, with no radiologist standing out.

Against the radiologist consensus, the model reached a QWK of 0.803 on the BMU Cohort and 0.823 on the Multi-District Cohort. Because the consensus combines three readers, it is a less noisy reference than any single radiologist, so these values are not directly comparable with inter-radiologist QWK; we report them as agreement with the reference standard rather than as evidence of radiologist-level grading. Disagreements with the consensus were nearly always by one grade: 5 of 271 BMU knees and 3 of 1,000 Multi-District knees differed by more {(Fig.~\ref{fig:performance}C,D)}. On the BMU Cohort, two of the five KL4 knees were identified, and both KL4 predictions were correct, but five knees are too few to judge this grade.

\section*{Discussion}

We set out to test whether a compact ordinal grader for knee osteoarthritis, developed on a public cohort from the United States, would hold up on radiographs from Bangladesh, and what was needed to adapt it when it did not. On the OAI test set, KneeFS-Net reached a QWK of 0.863 with about 0.8 million parameters. On the independent single-center BMU Cohort, its agreement with the radiologist consensus was 0.803. On radiographs from 219 centers in the Multi-District Cohort, direct transfer lowered agreement to 0.656, and self-supervised pretraining on unlabeled archive images followed by joint fine-tuning with a limited number of graded knees raised it to 0.823 in 5-fold cross-validation. These results show that a compact model can reach agreement comparable to much larger models on curated data, but that this agreement does not carry over to routine teleradiology images without adaptation to the target setting.

The main contributions of this work are:
\begin{itemize}
    \item KneeFS-Net, an ordinal grader with about 0.8 million parameters whose agreement on OAI is comparable to that of general-purpose models with 9 to 35 times more parameters.
    \item A measurement of how much agreement is lost under real-world domain shift, and a practical route to recover most of it using unlabeled archive images and a modest annotation effort.
    \item An external evaluation on two Bangladeshi cohorts, one single-center and one from 219 centers, in which three radiologists independently graded every labeled knee and the model was compared with each of them.
\end{itemize}

On OAI, KneeFS-Net reached agreement comparable to DenseNet121 and Swin-T at a fraction of their parameters, operations and memory (Table~\ref{tab:benchmark}). The low operation count follows from the design: dense connectivity with a growth width of 32 keeps each layer narrow, depthwise convolutions replace full convolutions, the Fourier branch mixes information across the whole feature map in $O(N \log N)$ operations rather than the $O(N^2)$ of self-attention, and the state-space scan, which is linear in sequence length, runs only on the $14\times14$ and $7\times7$ maps. Operation count does not translate directly into speed, however. The Fourier transforms and the sequential scan have low arithmetic intensity and run as many small GPU operations, so they are probably limited by memory traffic rather than computation, which explains why the model was slower per image than the larger comparators. Its advantage therefore lies in computation and memory, which matter most when a model runs on CPUs or low-memory hardware, rather than in GPU throughput; at under 10 ms per image, its latency remains well within what a reporting workflow can accept. We do not claim the best in-domain performance. Published models report higher agreement, including a QWK of 0.881 for KL-FuseNet on OAI \cite{alkhatatbehCrossInstitutionalFiveClassKellgren2026} and 0.902 for AGE-Net on a public KL dataset whose split was not reported \cite{liAGENetSpectralSpatialFusion2026}. Our re-implementations of these and two other published models scored below KneeFS-Net, but also below their published values, so that comparison says more about the difficulty of reproducing these models than about their true performance. The case for KneeFS-Net rests instead on its size and on how it behaves outside the development data. Compact designs have been proposed for knee radiographs before \cite{abdusalomovLightweightEarlyDetection2025}; here we evaluate one for full five-grade ordinal grading across two external cohorts.

On the BMU Cohort, the model's agreement with each radiologist (QWK 0.765--0.806; mean 0.784) was lower than the mean agreement between radiologists (0.849). The gap was larger on the Multi-District Cohort even after adaptation (mean model--radiologist QWK 0.764 vs.\ 0.869 between radiologists), although the radiologists agreed with one another at least as closely there as on the BMU Cohort. Noisier reference labels therefore do not explain it; the remaining shortfall more probably reflects image variation that the adapted model has not fully absorbed. On the BMU Cohort, the model's disagreements with the consensus were almost all between adjacent grades and were spread across KL0 to KL3 rather than concentrated at one cut-point (Fig.~\ref{fig:performance}C), which points to ambiguity at grade boundaries rather than a single miscalibrated threshold.

Generalization was the central concern of this study, and the two Bangladeshi cohorts behaved differently. Agreement fell only slightly on the BMU Cohort but substantially on the Multi-District Cohort, even though both had a higher proportion of KL1 knees than OAI. Differences in image appearance may explain this. On visual inspection, BMU radiographs looked much like OAI images. The Multi-District radiographs were mixed: some resembled OAI, but most were grainier and noisier, varied in scale, and often contained artifacts, consistent with the image-noise analysis (Fig.~\ref{fig:cohorts}). However, we did not test this explanation directly. The adaptation experiments suggest what labels were needed for. The label-free methods align the feature distributions of the two cohorts, CORAL by matching their second-order statistics and self-supervised pretraining by learning features from target-domain images, but neither can correct a change in how image features map to grades, such as a different reading convention or grade prevalence. Joint fine-tuning supplies this information directly, and most of the recovery came from it. Self-supervised initialization added comparatively little QWK but improved the macro-averaged metrics much more. QWK is dominated by the frequent mild grades and penalizes adjacent errors lightly, whereas macro-averaged metrics weight every grade equally, so the benefit of pretraining lay mainly in the less frequent grades, where features learned from 6,092 unlabeled target images reduce how many labeled examples the model needs. Alkhatatbeh et al. reached a similar conclusion in a Chinese cohort, where an OAI-trained model lost accuracy on transfer and recovered after fine-tuning on labeled local data \cite{alkhatatbehCrossInstitutionalFiveClassKellgren2026}. Our results add that a large unlabeled archive can be used alongside a limited set of graded knees. The evidence for self-supervised pretraining on musculoskeletal radiographs is mixed, however: domain-adapted DINO pretraining has also been reported to give no clear gain in classification and weak out-of-distribution performance \cite{bagchiPerformanceSelfSupervisedPretrained2026}. We therefore see self-supervised pretraining as a complement to target-domain labels rather than a substitute for them.

Class imbalance shaped both training and error patterns. In OAI, KL1 and KL4 make up 18.1\% and 3.0\% of knees, against 39.4\% for KL0. We addressed this with class-balanced sampling, stronger label smoothing and focal weighting for the less frequent and more ambiguous grades, extra weight on the KL1/KL2 and KL2/KL3 transitions, and an ordinal output in which these transitions are learned as separate cumulative outputs. When a model was trained jointly on OAI and Multi-District knees (Experiments E and F), we tuned a separate set of decision thresholds for each cohort on its own validation knees, because the grade distribution differed between them: KL0 was the most common grade in OAI and KL1 in the Multi-District Cohort. Models trained only with OAI labels, including the one evaluated on the BMU Cohort, used the OAI thresholds. Despite these measures, KL1 remained the hardest grade in OAI, and in the Multi-District Cohort it had the lowest recall of any grade (Fig.~\ref{fig:performance}A,D). This is expected, since KL1 is defined by doubtful joint-space narrowing and possible osteophytic lipping \cite{kellgren_radiological_1957}, and readers often disagree in distinguishing the early grades \cite{kohn_classifications_2016}. The direction of these errors differed between cohorts: on OAI, the model more often under-graded mild disease (KL1 as KL0), whereas on the Multi-District Cohort it more often over-graded (KL1 as KL2 and KL2 as KL3). A possible explanation is that the Multi-District thresholds, tuned on validation knees in which KL1 dominates, move the cut-points between the mild grades, although we did not test this directly. Severe disease was recognized reliably in the two larger cohorts: every KL4 knee in the OAI test set and all 23 KL4 knees in the Multi-District Cohort were identified. In the BMU Cohort, two of the five KL4 knees were identified, too few to judge this grade. For the final models, errors of more than one grade were rare in every cohort.

These results suggest a workable approach for settings like Bangladesh, where radiographs are plentiful but expert grading is scarce. A compact grader can be pretrained, or periodically re-adapted, on a teleradiology network's own unlabeled archive, with radiologists grading only a limited subset for fine-tuning and validation. The grading model itself is small, although the full pipeline also includes knee detection and region localization, whose cost we did not measure. We see the model as a triage tool that flags radiographs for prioritized expert review, not as a replacement for radiologist interpretation.

This study has several limitations. The region boxes of both cohorts of the BUET-KOA dataset were checked by experts and failed localizations were excluded, which a fully automated system could not do. All Multi-District images came from a single teleradiology provider, so the cohort does not represent Bangladesh, or South Asia, as a whole. The reference standards also differ: OAI grades come from central readers, whereas both Bangladeshi cohorts were graded by the same three local radiologists, whose habits may influence both the training labels and the evaluation. KL4 estimates rest on few knees, particularly in the BMU Cohort. The published comparison models were rebuilt from their descriptions and may not match the originals, and all results come from a single training seed. The comparison with other architectures on the Multi-District Cohort used a single 200-knee split rather than cross-validation. Finally, the study was retrospective, and no part of the pipeline has been tested in a prospective clinical workflow.

In conclusion, KneeFS-Net grades knee radiographs with agreement comparable to much larger models while using about 0.8 million parameters. On a Bangladeshi hospital cohort, it kept most of its agreement without retraining, although its agreement with individual radiologists remained below that between radiologists. Under the broader variation of routine teleradiology images, agreement fell, and was largely restored by combining self-supervised pretraining on unlabeled archive images with a limited number of graded knees. Prospective evaluation in clinical workflows is needed before the model is used to support patient care.

\section*{Methods}
\label{methods}
\subsection*{Study design}
 
We conducted a retrospective study using three cohorts of knee radiographs. The Osteoarthritis Initiative (OAI) served as the public source cohort for model development. Two Bangladeshi cohorts, which together form the BUET-KOA dataset, served as target cohorts: the BMU Cohort for external validation and the Multi-District Cohort for studying domain shift and target-domain adaptation.
 
OAI is a public, de-identified research dataset, and we used it under its data-use agreement. All BUET-KOA images were de-identified before analysis, and we used PaddleOCR to detect and remove embedded text and identifiers. The study involved no patient contact or intervention.

\subsection*{Study cohorts}

\begin{figure*}[htpb]
    \centering
    \includegraphics[width=\textwidth]{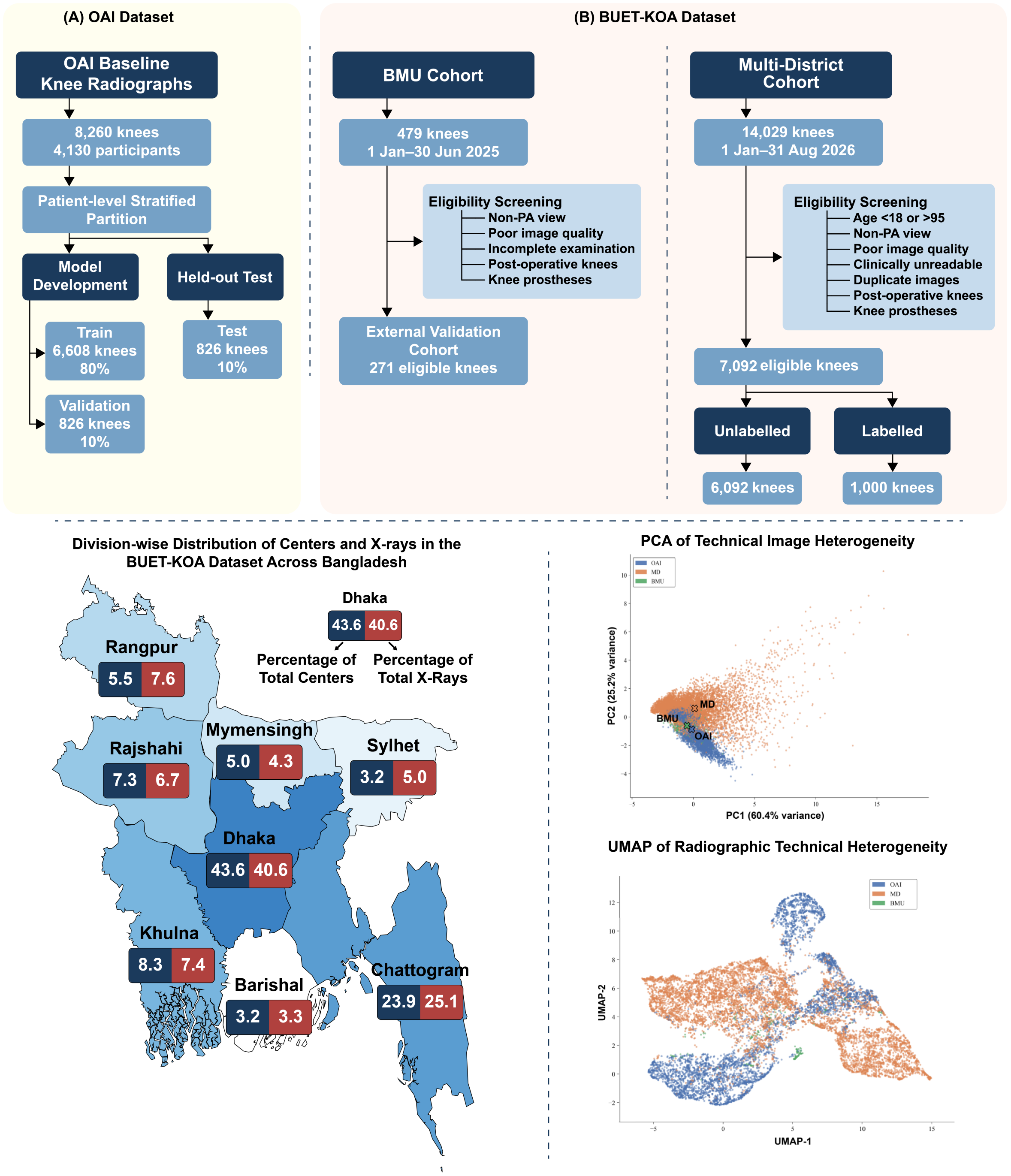}
    \caption{\textbf{Study cohorts.} \textbf{Top:} Selection and partitioning of the OAI (A) and BUET-KOA (B) datasets. \textbf{Bottom left:} Centers and radiographs of the Multi-District Cohort by administrative division. \textbf{Bottom right:} PCA and UMAP projections of image-noise descriptors.}
    \label{fig:cohorts}
\end{figure*}

Table~\ref{tab:demographics} summarizes the three cohorts, and Fig.~\ref{fig:cohorts} shows how each was selected and split.

\begin{table}[htpb]
\centering
\sffamily
\begin{threeparttable}
\caption{Demographic and clinical characteristics of the study cohorts}
\label{tab:demographics}
\setlength{\aboverulesep}{0pt}
\setlength{\belowrulesep}{0pt}
\renewcommand{\arraystretch}{1.25}
\begin{tabular}{l ccc}
\toprule[1.5pt]
\rowcolor{black!10}
\rule{0pt}{1.3em} & & \multicolumn{2}{c}{\textbf{BUET-KOA}} \\
\cmidrule(lr){3-4}
\rowcolor{black!10}
\textbf{Characteristic} & \textbf{OAI} & \textbf{BMU Cohort} & \textbf{Multi-District Cohort} \\[0.5ex]
\midrule[1pt]
\multicolumn{4}{l}{\textit{\textbf{Age, years}}} \\
Male & 64.11 $\pm$ 9.74 & 53.04 $\pm$ 5.67 & 39.56 $\pm$ 19.65 \\
Female & 64.56 $\pm$ 9.27 & 53.01 $\pm$ 5.36 & 45.95 $\pm$ 16.51 \\
Unspecified & --- & --- & 40.67 $\pm$ 18.32 \\
\textbf{Overall} & \textbf{64.37 $\pm$ 9.47} & \textbf{53.02 $\pm$ 5.50} & \textbf{42.86 $\pm$ 18.31} \\
\midrule
\multicolumn{4}{l}{\textit{\textbf{Sex, \%}}} \\
Male & 41.53 & 46.86 & 42.88 \\
Female & 58.46 & 53.14 & 50.49 \\
Unspecified & --- & --- & 6.63 \\
\midrule
\multicolumn{4}{l}{\textit{\textbf{KL grade, n}}} \\
KL0 & 3,253 & 71 & 170 \\
KL1 & 1,495 & 98 & 422 \\
KL2 & 2,175 & 73 & 310 \\
KL3 & 1,086 & 24 & 75 \\
KL4 & 251 & 5 & 23 \\
\textbf{n} & \textbf{8,260} & \textbf{271} & \textbf{1,000 (labeled)} \\
\bottomrule[1.5pt]
\end{tabular}
\begin{tablenotes}
    \item \footnotesize \textnormal{\textit{Note:} Values are mean $\pm$ standard deviation for age, percentages for sex, and counts for KL grade. Age and sex for the Multi-District Cohort are based on all eligible knees ($n$ = 7,092); KL grades are based on the 1,000 knees graded by radiologists and were unavailable for the remaining 6,092. The BMU and Multi-District Cohorts together form the BUET-KOA dataset.}
\end{tablenotes}
\end{threeparttable}
\end{table}

\subsubsection*{OAI}
\label{oai}

The OAI is a multicenter, longitudinal cohort study of knee osteoarthritis. We used the public release of 8,260 OAI knee images from 4,130 participants with Kellgren--Lawrence (KL) grades prepared by Chen et al.\ \cite{chen_knee_2018, Chen2019KneeOAOrdinalLoss}. We split the data at the participant level into training (80\%), validation (10\%) and test (10\%) subsets, stratified by KL grade, so that both knees of a participant always stayed in the same subset. Participant identifiers were recovered for splitting at patient level. The test subset was used only for the final evaluation.

\subsubsection*{BMU Cohort}
\label{bmu}

The BMU Cohort comes from Bangladesh Medical University, where knee radiographs acquired between 1 January and 30 June 2025 were collected through the RadAssist platform. Of the 479 radiographs in the archive, 271 knees remained after we excluded non-PA views, poor-quality or incomplete examinations, post-operative knees and knee prostheses. Every knee was graded by three radiologists. We never used this cohort to train or pretrain the grading model.

\subsubsection*{Multi-District Cohort}
\label{md}

The Multi-District Cohort comes from the archive of Telerad Bangladesh Ltd., a teleradiology network that reports radiographs for healthcare centers across Bangladesh. The archive held 14,029 knee radiographs acquired between 1 January and 31 August 2026 at 219 centers in multiple districts across all eight administrative divisions of the country (Fig.~\ref{fig:cohorts}, bottom left). Twenty-one radiologists had reported them in free text during routine practice, so no structured KL grades were available. We excluded patients younger than 18 or older than 95 years, non-PA views, poor-quality or clinically unreadable images, duplicates, post-operative knees and knee prostheses, leaving 7,092 knees.

We kept 6,092 of these knees unlabeled for self-supervised pretraining, and three radiologists graded the remaining 1,000. Each of the 1,000 labeled knees came from a different patient, so no patient contributed more than one knee to the labeled set. None of these patients had any image in the unlabeled pretraining pool. The labeled knees were divided into five folds of 200 knees, stratified by KL grade, which made the folds patient-level by construction. We evaluated every adaptation experiment with 5-fold cross-validation, so each labeled knee was used once as a test case. Because the centers are spread across the country, this cohort lets us study geographic and institutional shift within Bangladesh. It should not be read as representative of South Asia as a whole.
 
\subsection*{Reference standard}
\phantomsection\label{refstd}

Three radiologists graded every BMU knee and every labeled Multi-District knee independently. They were blinded to each other's grades, to the original reports and to clinical information, and graded using the standard KL definitions. The consensus grade was the majority vote; when all three grades differed, we used the grade of the senior radiologist. The same protocol was applied to both cohorts. Because each radiologist also contributed to the consensus, comparing the radiologists against it would favor them. We therefore kept the individual grades and compared the model with each radiologist, and each radiologist with each of the others, using the same agreement metric.

\subsection*{Image preprocessing and anatomical localization}
\label{preprocessing}
\label{landmark}
 
We converted all radiographs to 8-bit grayscale and applied the same preprocessing to pretraining, training and evaluation. Pixel intensities were scaled linearly to $[0,1]$ without histogram matching. Each knee crop was resized to the network input with bilinear interpolation, and region boxes were stored in normalized coordinates so that they remained valid at any input size.

For OAI, we located each knee using 74 landmarks along the femoral and tibial contours. Reference landmarks came from BoneFinder \cite{lindner_fully_2013}, and we trained a MobileNetV3-Large network with four landmark heads on OAI to predict them. Bilateral radiographs were split into single knees from the landmark bounding boxes, with a small margin to keep marginal osteophytes in view. We did not mirror the knees, because the regions were defined anatomically rather than by image position. From the landmarks, we defined three regions a priori: the medial edge, the lateral edge and the medial joint space. Each region box enclosed the relevant femoral and tibial landmarks, was padded by 35\% on each side, and was converted to binary masks at the $28\times28$ and $14\times14$ feature-map resolutions.

Landmark prediction was unreliable on BMU and Multi-District images, which vary widely in scale, view and quality, so both cohorts were processed through an identical pipeline that did not rely on landmarks. A YOLOv8 detector trained on OAI produced a square crop for each knee, and crops larger than the target were downsampled so that their size matched the distribution of OAI crops. An EfficientNet-B0 regressor, also trained on OAI, then predicted the three region boxes directly from each crop. The predicted boxes were converted to binary masks in the same way as for OAI and were checked visually by an expert before analysis.
 
\subsection*{KneeFS-Net}
\phantomsection\label{kneefsnet}

\begin{figure*}[!t]
    \centering
    \includegraphics[width=\textwidth]{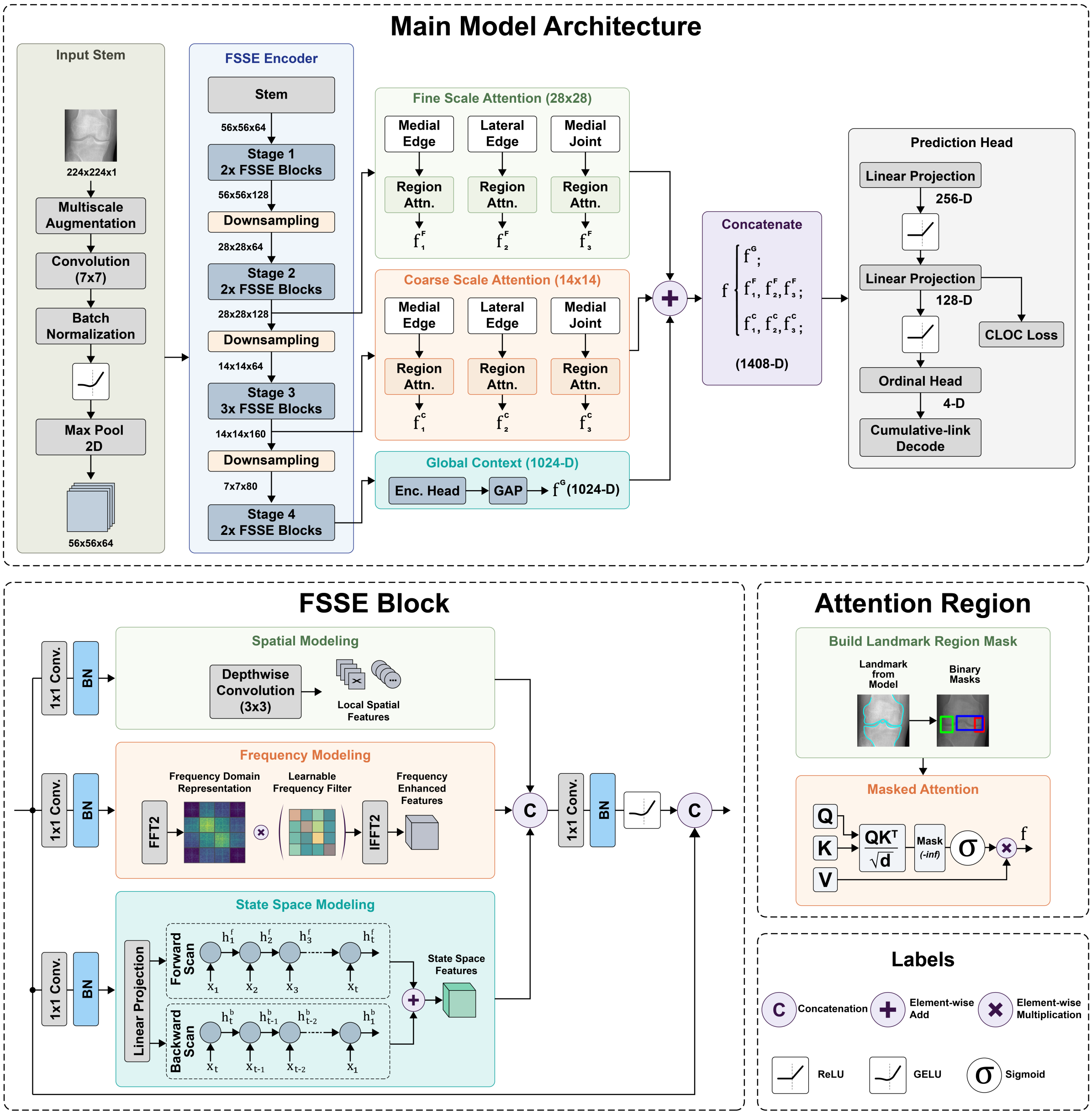}
    \caption{\textbf{KneeFS-Net architecture.} \textbf{Top:} Stem, four-stage FSSE encoder, two-scale region attention and ordinal head. \textbf{Bottom left:} FSSE layer; the state-space branch is used in stages 3 and 4 only. \textbf{Bottom right:} Masked region attention.}
    \label{fig:model_architecture}
\end{figure*}

KneeFS-Net takes a single-channel $224\times224$ knee image and predicts its KL grade (Fig.~\ref{fig:model_architecture}). It has three parts: an encoder built from frequency--spatial selective encoding (FSSE) layers, landmark-guided region attention at two scales, and an ordinal prediction head. The full model has 804,872 trainable parameters and needs 0.33 GFLOPs per image at this input size.

The stem is a $7\times7$ convolution with stride 2, followed by batch normalization, GELU and $3\times3$ max pooling, and turns the input into 64 feature channels at $56\times56$. Four stages follow, with 2, 2, 3 and 2 FSSE layers. Within a stage the layers are densely connected: each layer adds 32 new channels to everything computed before it. Between stages, a transition layer halves both the number of channels and the spatial resolution. The feature map therefore grows to 128 channels at $56\times56$ in stage 1, 128 channels at $28\times28$ in stage 2 and 160 channels at $14\times14$ in stage 3, and reaches 144 channels at $7\times7$ in stage 4. A final $1\times1$ convolution expands the stage-4 output to 1,024 channels. 

Each FSSE layer looks at its input $X_l$ through up to three branches. Each branch starts with a $1\times1$ convolution that reduces the input to the growth width of 32 channels. The local branch then applies a $3\times3$ depthwise convolution, which captures fine structure such as osteophyte margins and the edges of the joint space. The frequency branch computes
$Y_l = \mathcal{F}^{-1}\big(\mathcal{F}(X_l) \odot W_f\big)$,
where $\mathcal{F}$ is the two-dimensional Fourier transform and $W_f$ is a learnable complex-valued filter. The real and imaginary parts of $W_f$ are stored at a fixed size and interpolated to the frequency resolution of each feature map, so the same layer works at every scale. Because every frequency component mixes information from the whole map, this branch reaches across the knee at little cost.

In stages 3 and 4, a third branch flattens the feature map into a sequence of $L = H_l W_l$ positions and runs a selective state-space model,
$h_t = \bar{A}_t h_{t-1} + \bar{B}_t x_t$, $y_t = C_t h_t + D x_t$,
in which $\bar{A}_t$, $\bar{B}_t$ and $C_t$ depend on the input at each position. The scan runs forward and backward and the two outputs are summed. Because the hidden state is carried along the whole sequence, each output can draw on distant positions in the map, and the input-dependent parameters control how much information is passed on at each step. We use the branch only in the two deepest stages, where the maps are small ($14\times14$ and $7\times7$) and the scan adds little computation. The outputs of the active branches are concatenated, fused by a $1\times1$ convolution and appended to the layer input:
\begin{equation}
\label{eq:fsse}
    X_{l+1} = X_l \;\Vert\; \phi\!\left( W_c^{\,1\times1} \left[ F_{\mathrm{local}}(X_l) \Vert F_{\mathrm{freq}}(X_l) \Vert F_{\mathrm{ssm}}(X_l) \right] \right),
\end{equation}
where $\Vert$ denotes channel-wise concatenation, $W_c^{\,1\times1}$ is the fusion convolution and $\phi$ denotes batch normalization followed by GELU. In stages 1 and 2, $F_{\mathrm{ssm}}$ is omitted.

Global pooling of the encoder output describes the whole knee but can dilute small, localized changes. We therefore added region attention over the three regions defined from the landmarks (medial edge, lateral edge and medial joint space), applied to the stage-2 ($28\times28$) and stage-3 ($14\times14$) feature maps. For each region $r$, a learned 64-dimensional query $q_r$ attends only to positions inside the region mask:
\begin{equation}
\label{eq:attn}
    z_r = \operatorname{softmax}\!\left( \frac{q_r K^\top}{\sqrt{64}} + M_r \right) V,
\end{equation}
where $K$ and $V$ are $1\times1$ projections of the feature map, and $M_r$ is $0$ at positions inside the padded region box and $-\infty$ elsewhere. Positions outside the box therefore receive zero weight. The mask fixes where the model looks, and the learned query decides what inside each region matters. The finer scale keeps small structures such as early osteophytes visible, while the coarser scale summarizes each region with more context. Each $z_r$ is projected to 64 dimensions, so three regions at two scales give 384 regional features. We concatenate these with the 1,024-dimensional globally pooled encoder output to form a 1,408-dimensional representation of the knee.

A two-layer head (256 and 128 units, with dropout after the first layer) maps this representation to a 128-dimensional embedding $\mathbf{e}_i$, taken before the final activation and also used by the contrastive loss (see Supervised training). A linear layer then outputs four cumulative logits $z_{ik}$, one for each event $y_i > k$, $k \in \{0, \ldots, 3\}$. KL grades are ordered, and a prediction of KL3 for a KL2 knee is a smaller error than a prediction of KL4. The cumulative form builds this order into the output instead of treating KL0--KL4 as five unrelated classes. At inference, we averaged the logits of each image and its horizontal flip, $\bar{z}_{ik}$, and counted the thresholds crossed, $\hat{y}_i = \sum_{k} \mathbf{1}\big[\sigma(\bar{z}_{ik}) > \delta_k^{(d)}\big]$, where $\sigma$ is the logistic function and $\delta_k^{(d)}$ are thresholds tuned on the validation set of cohort $d$.
 
\subsection*{Self-supervised target-domain pretraining}
\label{ssl}
 
\begin{figure*}[t]
  \centering
  \includegraphics[width=\textwidth]{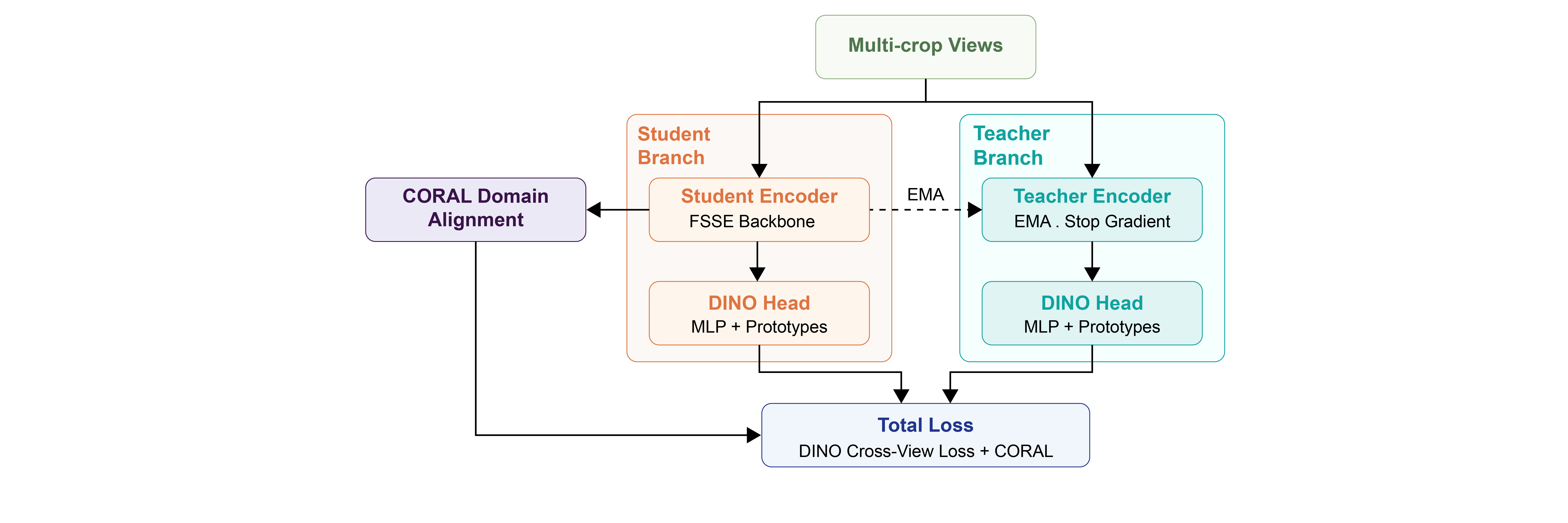}
    \caption{\textbf{Self-supervised pretraining of the FSSE encoder.} The student encodes two global and four local views; the teacher, updated as an exponential moving average of the student (dashed arrow), encodes the global views only. Teacher outputs are centered per cohort, and a CORAL penalty aligns student features across cohorts (Eq.~\eqref{eq:ssl-total}). The projection head is discarded after pretraining.}
  \label{fig:ssl}
\end{figure*}

To expose the encoder to Multi-District images without using their labels, we pretrained the FSSE encoder with DINO self-distillation \cite{caron_emerging_2021} (Fig.~\ref{fig:ssl}). The pretraining pool contained the OAI training split and the 6,092 unlabeled Multi-District knees. We left out the OAI validation and test splits, the 1,000 labeled Multi-District knees and all BMU images, and drew each batch equally from the two cohorts. Each image yielded two global views containing the medial joint space and four local views, three of them centered on the landmark regions. Matching local to global views therefore pushed the encoder to represent the same anatomy that the region attention uses later.
 
A pretext task like this can be solved by telling the cohorts apart rather than by learning anatomy, so we took four measures against it. The teacher outputs were centered separately for each cohort, so that the offset between cohort means could not become part of the target. Fourier domain adaptation \cite{yang_fda_2020} swapped low-frequency amplitude spectra between cohorts on randomly chosen views, and acquisition cues (resolution, sharpness, intensity, noise and compression) were randomized in both cohorts. Finally, a correlation alignment (CORAL) penalty \cite{sun_deep_2016} matched the means and covariances of the student's global features across cohorts. The full objective was
\begin{equation}
\label{eq:ssl-total}
\mathcal{L}_{\mathrm{SSL}} = \mathcal{L}_{\mathrm{DINO}} + w\,\mathcal{L}_{\mathrm{CORAL}} + \lambda_s\,\mathcal{L}_{\mathrm{spread}},
\end{equation}
where $\mathcal{L}_{\mathrm{DINO}}$ is the cross-view distillation loss of the original method \cite{caron_emerging_2021}, and $\mathcal{L}_{\mathrm{spread}}$ is a hinge penalty that keeps the spread of the normalized features within a fixed range, preventing both collapse and excessive dispersion.
 
We selected the checkpoint with a linear probe rather than the DINO loss, which keeps falling after the features stop improving. The probe was a logistic-regression classifier on frozen teacher features, trained with OAI training labels and scored by QWK on OAI validation; no KL labels entered the pretraining objective itself. The teacher encoder then initialized KneeFS-Net and the projection head was discarded, so pretraining adds no parameters at inference. Because the pretraining pool contained none of the labeled Multi-District knees, we pretrained the encoder once and used it in every cross-validation fold.
 
\subsection*{Supervised training and adaptation experiments}
\label{training}
 
We trained KneeFS-Net end to end on the cumulative targets $t_{ik} = \mathbf{1}(y_i > k)$ with
\begin{equation}
\label{eq:loss}
    \mathcal{L} = \mathcal{L}_{\mathrm{ord}} + 0.27\,\mathcal{L}_{\mathrm{focal}} + 0.2\,\mathcal{L}_{\mathrm{CLOC}}.
\end{equation}
$\mathcal{L}_{\mathrm{ord}}$ is a binary cross-entropy over the four targets, with stronger label smoothing for the ambiguous KL1 and KL2 grades, extra weight on the KL1/KL2 and KL2/KL3 transitions, and a penalty that keeps the cumulative logits monotonic. $\mathcal{L}_{\mathrm{focal}}$ is a focal loss whose focusing parameter depends on the true grade. $\mathcal{L}_{\mathrm{CLOC}}$ is a contrastive term inspired by CLOC (Contrastive Learning for Ordinal Classification) \cite{pitawelaCLOCContrastiveLearning2025}: it pulls embeddings of the same grade together and pushes different grades apart by a margin that grows with the grade difference. Unlike the multi-margin N-pair loss of the original work, we use a hinge form with four learnable adjacent margins.  We optimized with AdamW under a warm-up and cosine schedule, with class-balanced sampling and geometric and photometric augmentation applied jointly to each image and its region boxes.
 
We compared six training strategies with 5-fold cross-validation on the 1,000 labeled Multi-District knees (Table~\ref{tab:telerad_results}). Experiment A applied the OAI-trained model directly, and Experiment B trained only on the labeled Multi-District knees of the training folds. Experiments C and D used Multi-District images but not their labels: C applied CORAL (Correlation Alignment), aligning the covariance matrices of OAI and Multi-District feature representations to reduce domain discrepancy, and D started from the self-supervised encoder and was fine-tuned on OAI only. Experiments E and F trained jointly on OAI and the labeled knees of the training folds, with each batch split equally between the cohorts and thresholds tuned separately for each. F started from the self-supervised encoder and E from random weights, so the two differ only in initialization. In each fold, Experiments B, E and F were trained on four folds, with 200 of these knees held out as validation knees for model selection and threshold tuning, and tested on the remaining fold. Experiments A, C and D used no Multi-District labels, so the same model predicted all 1,000 knees. For every experiment, we pooled the predictions from the five test folds into one out-of-fold prediction per knee.

\subsection*{Comparator models}
\phantomsection\label{comparators}

All comparator models used the same data splits, input preprocessing and test set as KneeFS-Net. The four published KOA grading models were re-implemented from their publications or run from official code, where released. Where code was unavailable, we rebuilt each model from the details reported in its paper, so our versions may differ from the originals in unreported choices such as preprocessing, hyperparameters or training schedule, and their performance here may differ from the published figures. The baselines shared augmentation, tuning budget, threshold calibration, and test-time flipping with KneeFS-Net. For the Multi-District comparison, all comparator models received the same self-supervised pretraining and joint fine-tuning as Experiment F. This comparison, including the KneeFS-Net results reported with it, used a single stratified split of the labeled knees (600 training, 200 validation and 200 test) rather than cross-validation.

\subsection*{Statistical analysis}
 
QWK was the primary metric because it penalizes errors by their distance on the ordinal scale.    Secondary metrics were exact accuracy, within-one-grade agreement, mean absolute error, and macro-averaged F1, sensitivity and specificity; on the OAI test set we also report expected calibration error and the multiclass Brier score. We estimated 95\% confidence intervals by non-parametric bootstrap with 2,000 resamples of the test set; for the Multi-District Cohort, we resampled the pooled out-of-fold predictions of all 1,000 labeled knees, one per patient. We compared models, or the model and readers, with two-sided paired bootstrap tests on the same resamples. Two-sided $p$ values were computed as $2\min\{(n_{\le 0}+1)/(B+1),\,(n_{\ge 0}+1)/(B+1)\}$, where $n_{\le 0}$ and $n_{\ge 0}$ are the numbers of the $B = 2{,}000$ bootstrap differences at or below and at or above zero; the smallest attainable value is therefore 0.001. All results come from a single training run with a fixed seed in each fold. For the reader analysis, the mean QWK of the model with the individual radiologists was compared with the mean pairwise QWK between radiologists. The linear probe used to check the self-supervised encoder is described above.

\section*{Ethics Declaration}

The Osteoarthritis Initiative (OAI) dataset is a publicly available dataset and was used in accordance with its established data-use requirements. Our study was approved by the Institutional Review Board (IRB) of Bangladesh Medical University (BMU), Dhaka, Bangladesh (Ref: BMU/2026/2390 dated March 5, 2026). 
All BUET-KOA data were de-identified prior to analysis, and no directly identifiable patient information was used during this research. The study involved secondary analysis of existing radiographic data, with no patient recruitment or intervention.

\section*{Data and Code Availability}

The OAI dataset is publicly available. The code will be made publicly available upon acceptance. The Bangladeshi cohort data are available from the corresponding author on reasonable request.

\section*{Declaration of Competing Interests}
Y.R.T. is an employee of Telerad Bangladesh Ltd., which provided the radiographs of the Multi-District Cohort. Telerad Bangladesh Ltd. had no role in the study design, analysis, interpretation of results or the decision to publish. The remaining authors declare no competing interests.

\bibliographystyle{naturemag}
\bibliography{KneeFS}

\end{document}